# Lossless Image Compression Algorithm for Wireless Capsule Endoscopy by Content-Based Classification of Image Blocks

Atefe Rajaeefar, Ali Emami, S.M.Reza Soroushmehr,
Nader Karimi, Shadrokh Samavi, Kayvan Najarian

*Abstract*— Recent advances in capsule endoscopy systems have introduced new methods and capabilities. The capsule endoscopy system, by observing the entire digestive tract, has significantly improved diagnosing gastrointestinal disorders and diseases. The system has challenges such as the need to enhance the quality of the transmitted images, low frame rates of transmission, and battery lifetime that need to be addressed. One of the important parts of a capsule endoscopy system is the image compression unit. Better compression of images increases the frame rate and hence improves the diagnosis process. In this paper a high precision compression algorithm with high compression ratio is proposed. In this algorithm we use the similarity between frames to compress the data more efficiently.

## I. INTRODUCTION

Endoscopy process is used for medical diagnoses of various digestive diseases by observation of different parts of the digestive system such as esophagus, colon and small intestine, etc. Wired endoscopy is an uncomfortable and painful method due to using a flexible and long cable which is sent to the digestive tract; hence it is not suggested for young people [1]. Wireless endoscopy capsule (WCE) was introduced by Given Imaging Incorporated [2]. WCE may have some significant errors due to limited field of view, lower resolution and lower frame rates in comparison to push endoscopy [3].

Generally, an endoscopic capsule system is composed of three major parts: the ingestible capsule, a portable image recording belt or jacket, and a workstation computer with image processing software. Electronic ingestible capsule uses a CMOS camera which takes pictures of the tract and compresses the captured images. The capsule then sends the images to a radio frequency (RF) transmitter for sending them to a receiver which is out of the body [4]. A typical capsule and its receiver are shown in Fig. 1. This capsule requires a battery that should work for more than 16 hours. Hence, the electronic system should have low power consumption. Another problem is the low resolution of captured images which is usually insufficient for medical diagnosis and hence the image resolution should be enhanced. One solution for increasing the visual quality of the captured frames is to increase the frame rate. However, increasing image resolution and frame rate lead to significant increase in power consumption during radio frequency transmission of the images [1]. In order to overcome these issues, a strong compression algorithm is essential to reduce the size of transmitted images without losing the quality of the captured sequence.

WCE uses a CMOS image sensor with a Bayer color filter array (CFA) for capturing color images of the gastrointestinal (GI) tract. The captured CFA image consists of 2×2 repeating patterns with two green (G), one red (R), and one blue (B) pixels. Therefore, to produce full-color image, the two missing color components at each pixel location of the CFA should be recovered by a method called demosaicing [13]. Although the generated images have low inter-pixel correlation, using prediction based methods would not result in high performance. Some methods, such as [3], have attempted to solve this problem by separating pixels of the same color.

In light of the above discussion, compression unit is arguably a critical part of the endoscopic capsule system, for which different methods have been proposed. The suggested approaches may be divided into two main categories.

*Category I* belongs to lossless and near lossless methods such as JPEG-LS based methods [5-8], which provide proper image quality but achieve low compression ratio (CR) and low frame rate [3]. Compression ratio is defined as the ratio

A. Rajaeefar and N. Karimi, are with the Department of Electrical and Computer Engineering, Isfahan University of Technology, Isfahan 84156-83111, Iran.

A. Emami is with the Department of Electrical and Computer Engineering, Isfahan University of Technology, Isfahan 84156-83111, Iran. He is also with the Dept. of Information Technology and Elect. Engineering, University of Queensland, Brisbane, Australia.

S.M. R. Soroushmehr is with the Department of Computational Medicine and Bioinformatics and Michigan Center for Integrative Research in Critical Care, University of Michigan, Ann Arbor, MI, U.S.A.

S. Samavi is with the Department of Electrical and Computer Engineering, Isfahan University of Technology, Isfahan 84156-83111, Iran. He is also with the Department of Emergency Medicine, University of Michigan, Ann Arbor, MI, U.S.A.

Kayvan Najarian is with the Department of Computational Medicine and Bioinformatics, Department of Emergency Medicine and the Michigan Center for Integrative Research in Critical Care, University of Michigan, Ann Arbor, MI, U.S.A.

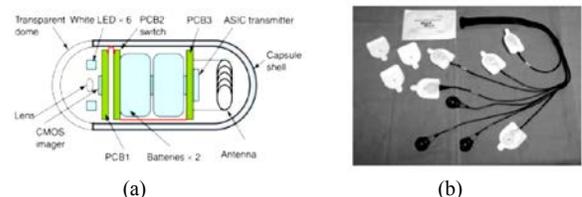

Fig. 1. (a) Shematic of capsule endoscopy system, (b) outside receiver sensors [1].

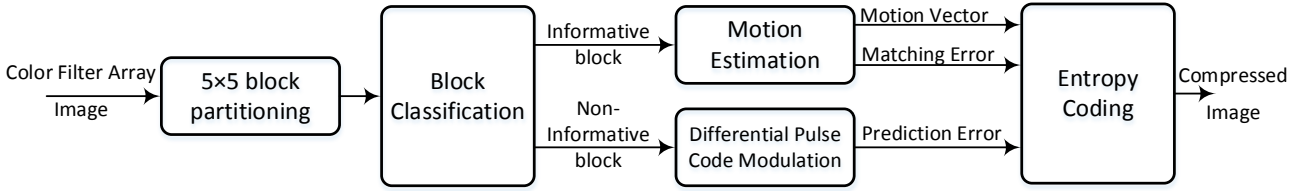

Fig. 2. Block diagram of proposed image compression system for WCE.

of the size of uncompressed image, $S_{in}$, to the size of the compressed image $S_{out}$. This is defined by Equation (1).

$$CR = \frac{S_{in}}{S_{out}} \qquad (1)$$

Among such lossless methods, the best compression rate is achieved by Xie et al. [8] which is 3.57.

*Category II*: The second category of compression methods, which are used for compression of WCE images, belongs to lossy algorithms. These lossy methods are mainly based on discrete cosine transform (DCT). Examples of lossy methods can be found in [9-13] which provide high compression rates. Despite their high performance in terms of compression ratios, these methods, according to [1], generate low quality images. These lossy methods provide high frame rates but generating of high resolution of images is still an issue in such methods.

As there is low correlation between neighboring pixels in CFA images, JPEG based algorithms do not perform well. Thus, we need modified versions of JPEG-based methods. Turcaze et al. [3] propose a near-lossless compression, which uses DCT transform on pixels of the same group. These groups consist of same color pixels. Then for the DC coefficients of the transform, their algorithm performs a Differential Pulse-Code Modulation (DPCM), and for the AC coefficients, the algorithm conducts quantization [3]. Eventually, the algorithm encodes all of the results with a customized entropy coder. The method of [3] increases the number of frames, which leads to increase of transmitted data and bit-rate. Furthermore, the quantization of DCT coefficients causes diagnosis errors.

In this paper, we introduce a lossless compression method with high compression ratio. . By increasing the frame rate, the amount of transmitted data per frame is decreased.

In the rest of the paper, we elaborate on different parts of the algorithm in section II and experimental results are presented in section III. Concluding remarks are offered in section IV of the paper.

## II. PROPOSED ALGORITHM

In order to increase the accuracy of diagnosis, we increase the number of transmitted frames. Because in low frame rates, some parts of the tract might be missed. However, this approach leads to the increase of data transmission. Due to limitation of power resource, we are limited in data transmission bit-rate. To alleviate this problem, in this section we propose a compression method based on similarity of consecutive frames rather than taking advantage of spatial similarity. In this way, increasing the frame rate does not significantly increase data transmission rate. The block diagram of the proposed algorithm is shown in Fig. 2. In most capsule endoscopy systems, real time data compression is not possible and some of the images are stored on a chip memory. However, in our proposal, increasing the frame rate leads to higher similarity among consecutive frames. This idea has been previously used for video compression. Here, we propose an optimization of the algorithm for capsule endoscopic images.

### A. Blocking module

As WCE uses CFA images, first we should change the colors of the images in the database then insert them to the module. Hence, for each pixel we keep one channel and throw away other channels.

To compare multiple frames, we divide each frame into blocks. The size of the blocks depends on the size of the image, edges, patterns of the image, and degree of similarity between frames. The best size of the block can be found by iteratively searching for the optimum size.

### B. Block Classification

Searching all blocks is a time and area consuming part which is not necessary for all blocks. In endoscopy images there are some dark and smooth non-informative parts. For these parts we calculate a gradient for each block. In this method gradient is calculated by a simple 3×3 Sobel filter. Two Sobel filters for computing the gradients in the $x$ and $y$ directions, as $G_x$ and $G_y$, shown in Fig. 3. The absolute values of these values are compared with a predefined threshold. If the gradient of a block is below the threshold, the block is labeled as non-informative (smooth) and its data is compressed using DPCM. Otherwise, motion estimation is performed for the block. To mark smooth blocks, and identify them in the receiver, we set their motion-vector (MV) values ($x$ and $y$) to 4 and 0.

### C. Motion Estimation Module

In the search module, the first frame is compressed directly with JPEG-LS. The second frame is divided to blocks and for each block in the current frame, we look for the best block that matches the current block in a search area of the previous frame. For choosing the best block, the mean square error (MSE) between the current block and every block in the search area is calculated and the block with the minimum MSE is selected. Then, we subtract these two blocks (i.e. selected and the current blocks) and also store the $x$ and $y$ positions of the selected block in an MV array.

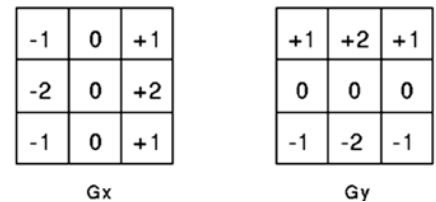

Fig. 3. Sobel filters for computation of gradients in $x$ and $y$ directions.

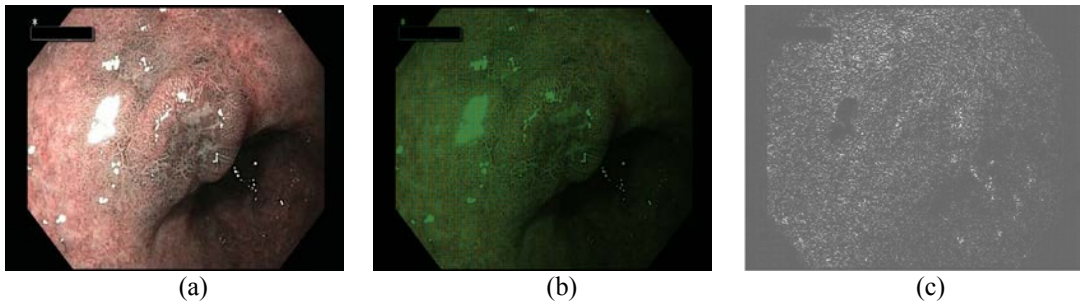

(a)           (b)           (c)

Fig. 4. (a) A captured WCE image [14], (b) CFA filter array image, (c) error image before Golomb Rice entroy encoding.

### D. Prediction Module

For smooth (non-informative) blocks, intensities of pixels are close to each other and the difference between adjacent pixels is low. Hence, for smooth blocks DPCM is an efficient compression method. The first pixel of the block is saved and then the difference of current pixel and the next pixel is calculated as the prediction error. Computation of the differences among adjacent pixels could be performed in parallel which could significantly decrease the execution time of this module.

### E. Entropy Coder

For removing statistical redundancy we can use a separate JPEG or JPEG-LS block to compress the error data. However, for endoscopy images such an elaborate module would not be feasible. Instead, a simple Golomb Rice entropy decoder is sufficient. A customized Golomb Rice algorithm is discussed in [3].

The Golomb-Rice code word of an integer $x \geq 0$ consists of two parts: a unary representation of $x/2k$ followed by the $k$ least significant bits of $x$, where $k \geq 0$ is an integer parameter of the code. In this algorithm, first we should convert signed input data to unsigned data and then choose an optimized $k$ parameter for best encoding results. The parameter $k$ should follow the local statistics of data. Hence, $k$ is adaptively changed with input data.

## III. EXPERIMENTAL RESULT

For evaluating the algorithm, we use two databases: one is the video database in the YouTube page of Given Imaging Inc. [14] and another one is push endoscopy images [15]. For the first database, we extract 10 frames per second from the video. After extracting them, we convert them to CFA images. Input image, color filtered and error image of capsule endoscopy are shown in Fig. 4. Input image, color filtered and error image of push endoscopy images are shown in Fig. 5.

### A. Algorithm parameters

In this algorithm the threshold value for selecting smooth (non-informative) blocks is 10. The block size is chosen 5×5 due to the size of the images in database [14] which is 480×480. An optimal search area for finding the best matched block is $2 \times N + 1$, where $N$ is the length of the block. Hence, the search area in this experiment is 11×11 pixels.

### A. Results

To evaluate the compression efficiency, we use compression ratio.

Compression ratio is calculated by considering the size of both errors and motion vectors, as shown in the block diagram of Fig. 2. Although motion vectors are extra data generated for compression algorithm, their size is not comparable to the size of error images, as there is only one MV per each block.

For 40 frames of the first database [14] the amount of compression ratio (CR) without removing spatial redundancy and by using smooth module is 9.75 whereas using JPEG-Ls on errors results in the CR of 10.93. Without using smooth module, the amount of CR without using JPEG-LS is 10.73 and with JPEG-LS, CR becomes 12.66. In comparison to other methods, for capsule endoscopy image database, the method proposed in [3] achieves CR of 3.94 and the method proposed in [16] achieves CR of 3.88. Also using JPEG-LS method gives a CR of 1.98 which is shown in Table 1. The CR of the proposed method helps decrease the transmission rate and allows increase of the frame rate, which makes the resulting imaging more trustable.

For 40 images of the second database [15] which have higher resolution, CR is 2.5 and for JPEG-LS the CR of 1.66

TABLE 1. Average compression ratio for capsule endoscopic images extracted from videos of [14] compared with other methods.

| Compression Ratio | | | |
|---|---|---|---|
| Proposed Algorithm | Turcaze et al [3] | Chen et al [16] | JPEG-LS |
| **10.93** | 3.94 | 3.88 | 1.98 |

TABLE 2. Average compression ratio for push endoscopy images of [15] compared with JPEG-LS.

| Compression Ratio | |
|---|---|
| Proposed Algorithm | JPEG-LS |
| **2.50** | 1.66 |

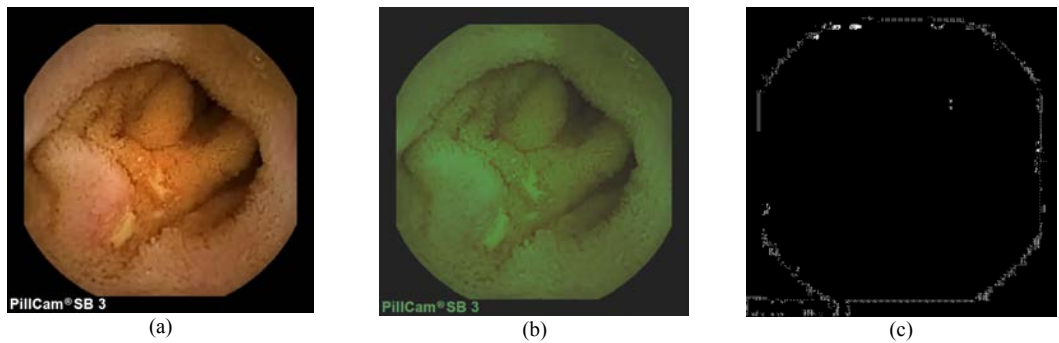

Fig. 5. (a) Captured capsule endoscopy images [15], (b) CFA filter array image, (c) error image before entropy encoder.

is achieved, as shown in Table 2. As the size and resolution of images increase, performance of the block matching decreases. However, the proposed method has higher CR and also its frame rate is higher than other methods.

Complete implementation of this algorithm on the hardware is a bit more complicated than other comparable methods. With the advances that are occurring in the field of Very Large Scale Integrated circuits (VLSI), implementation of the proposed method is possible and the circuit area will be small enough for the WCE systems. By simplifying the method and transforming to near-lossless the amount of CR will slightly decrease with almost no change in the performance of the compression algorithm.

## IV. CONCLUSION

In this paper we proposed a compression method which resulted in relatively high compression ratios, for wireless capsule endoscopy systems. Simulations of the proposed method proved its good performance. To perform full search, we need to compare a current block with 49 reference blocks. The design goal was simplicity and high performance. By using CFA images and searching each color in its appropriate region, we can reduce the number of comparisons from 49 to only 9. Also by optimizing different parts of the algorithm a near lossless method can be obtained. For example, by quantization of the prediction errors and reduction of the number of block searches in a search area we can reduce the required hardware and reduce the power consumption of the system. Furthermore, size of the blocks and size of the search area can be considered as variables for reduction of hardware and power usage.

With the current trends in advancement of VLSI technology we expect reduction of transistor size and reduction of required supply voltage. Hence, implementation of the proposed design in current or near future capsule endoscopy systems is very feasible.

## REFERENCES

[1] Alam, M. W., Hasan, M. M., Mohammed, S. K., Deeba, F., & Wahid, K. A. (2017). Are Current Advances of Compression Algorithms for Capsule Endoscopy Enough? A Technical Review. *IEEE reviews in biomedical engineering*, 2017.

[2] Iddan, G., Meron, G., Glukhovsky, A. and Swain, P., "Wireless capsule endoscopy," *Nature*, 405(6785), p.417, 2000.

[3] P. Turcza and M. Duplaga, "Near-lossless energy-efficient image compression algorithm for wireless capsule endoscopy," *Biomed. Signal Process. Control*, vol. 38, pp. 1–8, 2017.

[4] M. R. Basar, F. Malek, K. M. Juni, M. S. Idris, and M. I. M. Saleh, "Ingestible Wireless Capsule Technology: A Review of Development and Future Indication," *Int. J. Antennas Propagation*, vol. 2012, pp. 1–14, 2012.

[5] X. Xie, G. Li, X. Chen, L. Liu, C. Zhang, and Z. Wang, "A Low Power Digital IC Design Inside the Wireless Endoscopy Capsule," in *IEEE Asian Solid-State Circuits Conference*, no. 60372021, pp. 217–220, 2005.

[6] G. Liu, G. Yan, S. Zhao, and S. Kuang, "A complexity-efficient and one-pass image compression algorithm for wireless capsule endoscopy," *Technol. Heal. Care*, vol. 23, no. s2, pp. S239–S247, 2015.

[7] C. G. Liu, G. Yan, B. Zhu, and L. Lu, "Design of a video capsule endoscopy system with low-power ASIC for monitoring gastrointestinal tract," *Med. Biol. Eng. Comput.*, vol. 54, no. 11, pp. 1779–1791, 2016.

[8] X. Xiang, L. GuoLin, C. XinKai, L. Lu, Z. Chun, and W. ZhiHua, "A low power digital IC design inside the wireless endoscopy capsule," *IEEE Asian Solid-State Circuits Conf. ASSCC*, vol. 41, no. 11, pp. 217–220, 2006.

[9] P. Turcza and M. Duplaga, "Hardware-efficient low-power image processing system for wireless capsule endoscopy," *IEEE J. Biomed. Heal. informatics*, vol. 17, no. 6, pp. 1046–56, Nov. 2013.

[10] J. Li and Y. Deng, "Fast Compression Algorithms for Capsule Endoscope Images," in *2009 2nd International Congress on Image and Signal Processing*, pp. 1–4, 2009.

[11] Y. Gu, X. Xie, Z. Wang, G. Li, and T. Sun, "Two-stage wireless capsule image compression with low complexity and high quality," *Electron. Lett.*, vol. 48, no. 25, pp. 1588–1589, 2012.

[12] Khan Wahid, Seok-Bum Ko, and D. Teng, "Efficient hardware implementation of an image compressor for wireless capsule endoscopy applications," in *IEEE International Joint Conference on Neural Networks*, pp. 2761–2765, 2008.

[13] H.S. Malvar, L.W. He, R. Cutler, "High-quality linear interpolation for demosaicing of bayer-patterned color images," in: Proc. ICASSP, vol. 3, pp. 485–488, 2004

[14] Given Imaging YouTube channel: https://www.youtube.com/user/GivenImagingIntl/videos.

[15] M. Ye, S. Giannarou, A. Meining, G.-Z. Yang. "Online Tracking and Retargeting with Applications to Optical Biopsy in Gastrointestinal Endoscopic Examinations". *Medical Image Analysis*. 2015.

[16] X. Chen, X. Zhang, L. Zhang, X. Li, N. Qi, H. Jiang, Z. Wang, "A wireless capsule endoscope system with low power controlling and processing", ASIC, IEEE Trans. Biomed. Circuits Syst. 3 (1) (Feb 2009) 11–22.